\relax
\documentclass[twoside,leqno,twocolumn]{article}
\usepackage{ltexpprt}  %Required
\usepackage{times}  %Required
\usepackage{helvet}  %Required
\usepackage{courier}  %Required
\usepackage{url}  %Required
\usepackage{graphicx}  %Required
\usepackage[square,sort,comma,numbers]{natbib}
\usepackage{graphicx,color}
\usepackage{amssymb,amsmath}
\usepackage{natbib}
\usepackage{algorithm}
\usepackage{algorithmic}
\usepackage[algo2e]{algorithm2e}
\usepackage{subcaption}
\usepackage{array}
\usepackage{enumitem,calc,flushend}

\newcommand{\norm}[1]{\left\lVert#1\right\rVert}
\newcommand{\hide}[1]{}

\begin{document}

% \title{Fully Adaptive Deep Model for Multi-task Multi-view Learning}
\title{Deep Multimodality Model for Multi-task Multi-view Learning}
\author{ Lecheng Zheng \thanks{Arizona State University.Email: $\{$lzheng43, jingrui.he$\}$@asu.edu}~~~Yu Cheng \thanks{Microsoft AI \& Research.Email: yu.cheng@microsoft.com}~~~Jingrui He$^*$}

\date{}
\maketitle
\begin{abstract}
  Many real-world problems exhibit the coexistence of multiple types of heterogeneity, such as view heterogeneity (i.e., multi-view property) and task heterogeneity (i.e., multi-task property). For example, in an image classification problem containing multiple poses of the same object, each pose can be considered as one view, and the detection of each type of object can be treated as one task. Furthermore, in some problems, the data type of multiple views might be different. In a web classification problem, for instance, we might be provided an image and text mixed data set, where the web pages are characterized by both images and texts. A common strategy to solve this kind of problem is to leverage the consistency of views and the relatedness of tasks to build the prediction model. In the context of deep neural network, multi-task relatedness is usually realized by grouping tasks at each layer, while multi-view consistency is usually enforced by finding the maximal correlation coefficient between views. However, there is no existing deep learning algorithm that jointly models task and view dual heterogeneity, particularly for a data set with multiple modalities (text and image mixed data set or text and video mixed data set, etc.). In this paper, we bridge this gap by proposing a deep multi-task multi-view learning framework that learns a deep representation for such dual-heterogeneity problems. Empirical studies on multiple real-world data sets demonstrate the effectiveness of our proposed Deep-MTMV algorithm.
\end{abstract}
\\
\textbf{Keywords:}\\
    Multi-view Learning, Multi-task Learning, Deep learning

\section{Introduction}
In contrast to the single view or single task in a traditional classification setting, it is usually the case that many real-world  problems have multiple views or multiple tasks or both of them. For example, in web classification problems, each web page can be characterized by multiple sources, including web title, web links, content in the web, etc. Each source can be considered as one view and usually contains the complementary information to each other. In image classification problems, the classifiers could learn to distinguish the domestic animals from the wild animals and also to classify the object in the image to be a cat or a dog, while the different views could be the distinct poses of the same animal.

% Another example is document classification, where given the first classification task of Tweets and the second classification task of conventional documents, they share the common view such as the TF-IDF representation; but at the same time, they have their own task-specific views, such as the social network for twitter users. 
% In particular, some data collected from Twitter contain not only the text but also the images surrounded by it.

Up until now, some researchers have proposed a variety of techniques to model a single type of heterogeneity. For example, in multi-view learning, \cite{yu2011bayesian} proposed an undirected graphical model to minimize the disagreement between multi-view classifiers; \cite{ZhouH17} follows the principles of view consistency by regularizing the prediction tensor.
in multi-task learning, the intuition is that tasks usually share the same structure, such as the tree structure in \cite{kim2010tree}, the clustered structure in \cite{jacob2009clustered}
% , or the graph-guided structure in \cite{chen2010graph}
, etc. However, for the real-world problems that exhibit view and task dual heterogeneity, only making use of the techniques from multi-view learning or from multi-task learning is not able to achieve the optimal performance. To address this problem, \cite{he2011graphbased} proposed a graph-based framework for multi-task multi-view learning (M2TV) that models both types of heterogeneity to help classify the unlabeled data. 
% \cite{hong2013tracking} imposed the L1,2-norm group-sparsity regularization to exploit the inter-dependencies among different views, and decomposed the sparse representation into two matrices to find the relatedness of tasks and to detect outlier tasks simultaneously. 
\cite{lu2017multilinear} proposed multilinear factorization machines, which can capture the relationships between multiple tasks with multiple views by constructing the task-view shared multilinear structures and learn the task-specific feature map.
Despite the fact that these algorithms can deal with textual data very well, they fail to capture the spatial information of image data by just vectorizing the images. 
 
% Recently, convolutional neural network is widely used for image data set because of its ability to preserve the spatial information of image. Deep learning becomes popular due to its ability to extract important features automatically, its effectiveness for very large data set and the state-of-the-art results. In deep multi-view learning, the authors of \cite{chang2015heterogeneous} showed that a common feature representation of different views can be created by minimizing the loss in this unified feature space. In deep multi-task learning, a cross-stitch network proposed in \cite{misra2016cross} aimed to find the relatedness of two tasks in almost each hidden layer. Some of these algorithms explore the relatedness of different tasks, but they disregard the view consistency of a single task; whereas the existing multi-view deep learning algorithm also fails to explore to the relatedness of different tasks. In addition, these multi-task or multi-view models can only deal with either image data or text data, not to mention a mixture of image and text data set. To the best of our knowledge, there does not exist a multi-view multi-task learning algorithm in context of deep neural network to fully utilize the view heterogeneity and task heterogeneity concurrently and to cope with text-image data set simultaneously.

Recently, deep learning techniques have been successfully applied to model various types of data, such as image data \cite{lu2016fully,liu2016deepfashion} and text data \cite{kim2014convolutional,misra2016cross} with significantly improved performance and important features extracted in an automatic way. For example, in deep multi-view learning, the authors of \cite{chang2015heterogeneous} showed that a common feature representation of different views can be created by minimizing the loss in this unified feature space; in deep multi-task learning, a cross-stitch network proposed in \cite{misra2016cross} aimed to find the relatedness of two tasks in almost each hidden layer. However, to the best of our knowledge, there does not exist a deep learning algorithm for modeling view and task dual heterogeneity. In other words, existing deep neural network structures only take into consideration task or view heterogeneity, and cannot be naturally extended to model both types. 
% Furthermore, existing multi-task or multi-view deep learning models can only deal with a single type of data, and cannot be used on multi-modality data.

To bridge this gap, we propose a deep multi-task multi-view learning framework that can model multi-modality data. The key idea is that for different views, we construct a different neural network with one unit per layer at the beginning based on the data type (see Figure \ref{fig1} for the architecture of the proposed model), and the complementary and consensus principles between these views are enforced by adding a regularization layer to constrain the output of multiple neural networks to be consistent. To integrate the output of these neural networks for multi-modality data, the weight of each view is automatically learned in the regularization layer and these weights are used to measure the contribution of each view to the final output. For different tasks, we group related tasks or attribute classifiers starting from the output layer to the input layer based on the similarity among tasks. Combining these two aspects, we propose an iterative algorithm to obtain the optimal estimates of the model parameters.
Our main contributions are summarized below: 
\begin{itemize}
    \item A novel deep heterogeneous learning framework addressing task and view dual heterogeneity;
    \item A generalized deep learning framework for modeling multi-modality data;
    \item A regularization layer designed to maximize the consistency of multiple views;
    \item Experimental results on several data sets demonstrating the effectiveness of the proposed framework.
\end{itemize}

The rest of this paper is organized as follows. A brief review of the related work is discussed in Section 2. Then we introduce our proposed framework for deep multi-view multi-task learning in Section 3. In Section 4, we evaluate our framework on multiple data sets. Finally, we conclude the paper in Section 5.

\section{Related Work}
In this section, we briefly review the related work on multi-view learning, multi-task learning, multi-view multi-task learning, as well as convolutional neural network (CNN).

\subsection{Multi-view Learning}
Multi-view learning has been studied for decades.~\cite{sindhwani2005co} proposed Co-regularization method to jointly regularize two Reproducing Kernel Hilbert Space $H^1$ and $H^2$. ~\cite{andrew2013deep} proposed Deep Canonical Correlation Analysis, which aims to find two deep networks such that the output layers of two networks are maximally correlated. In addition, multi-view Clustering (MVC) is another popular method used in unsupervised and semi-supervised learning and it aims to find several clusters such that similar data points are assigned to the same cluster and dissimilar data points are assigned to the different cluster by combining information from multiple views. \cite{kumar2011co} proposed a co-regularized multi-view clustering method by minimizing the disagreement between any pair of views. 
In this paper, we consider different tasks or attributes classifiers as data points and group these tasks by implementing multi-view clustering approach based on the similarities between tasks. 

% \vspace{-0.1cm}
\subsection{Multi-task Learning}
In parameter-based multi-task learning, task clustering approach and task relation learning approach are the most common strategies used to group tasks \cite{zhang2017survey,wangCF16}. The authors of \cite{zhou2011clustered} proposed a multi-task learning algorithm called CMTL, which assumes that each task can learn equally well from any other task.
In feature-based multi-task learning, it assumes that different tasks share the same feature representation derived from the original feature under the regularization framework \cite{argyriou2007multi}. In deep multi-task learning, \cite{lu2016fully} used top-down layer-wise widening method to split one unit layer into several branches and group tasks in this layer based on the affinity of tasks. 
\cite{zhang2014facial} proposed the tasks-constrained deep convolutional network method to formulate a task-constrained loss function, back-propagate the errors of related tasks jointly, and thus, improve the generalization of landmark detection.
In our paper, we combine the relatedness of tasks from multiple views to determine how tasks are clustered in a more precise way. 
% \yuc{Add more work on deep multitask learning, since your current work is derived from MTK. Done}

\subsection{Multi-view Multi-task Learning}
To cope with some real-world problems involving multiple views and multiple tasks, some researchers proposed to jointly model the two types of data heterogeneity. For example, \cite{liu2016urban} proposed spatio-temporal multi-task multi-view learning framework to predict the urban water quality, which fuses the heterogeneous data by penalizing the disagreement among different views and capture the spatial correlation among tasks by a graph Laplacian regularizer; \cite{sdmZhouYH17} seeks to find a weight tensor to represent the worker's behaviors across multiple tasks by exploiting the structured information. \cite{jin2014multi} proposed a method to learn the feature transformation for different views by classical linear discriminant analysis \cite{fukunaga2013introduction} and explore the shared task-specific structure for different tasks. However, most of these methods are only good at dealing with one type of data and they might deteriorate with another type of data. For example, the approach proposed in~\cite{he2011graphbased} has good performance for text data but it ignores the spatial information of image data by just vectorizing the images. In contrast, in this paper, we construct different types of neural networks for distinct data types, utilize the complementary information among different views, and exploit the relatedness of tasks to improve the performance of our proposed method.

\subsection{Convolutional Neural Network}
Two types of CNN are widely used for two types of data, including two-dimensional CNN for image data and one-dimensional CNN for text data. VGG-16~\cite{simonyan2014very} is one of the most famous two-dimensional CNN (2d CNN) architectures that are widely used to solve image classification problems. Different from two-dimensional CNN, ~\cite{kim2014convolutional} proposed a model based on one-dimensional convolutional neural network (1d CNN) for sentence classification. At first,  word2vec~\cite{mikolov2013distributed} is applied to find word embedding and each word has its own word vector in the feature space with dimensionality $R^{1 \times d}$. Then, each documentation can form a matrix $M \in R^{k \times d}$ by concatenating words in the documentation together, where $k$ is the maximal number of words in all documentations. N-grams are realized by training the different sizes of kernels. For example, a kernel $K \in R^{2 \times d}$ can extract a bi-gram. In this paper, we use 1d CNN to extract N-grams from text data and 2d CNN for image data.

\section{The Proposed Deep Multi-view Multi-task Learning}
In this section, we introduce our proposed deep framework for multi-view multi-task learning , which is able to simultaneously address multi-modality data.

\subsection{Preliminaries}

In this subsection, we briefly review the existing work of~\cite{lu2016fully}, which paves the way to our proposed framework. More specifically, the authors of~\cite{lu2016fully} proposed an adaptive layer-wise widening model to automatically learn a multi-task architecture based on a thin version of the VGG-16 network~\cite{simonyan2014very}. The core procedure is to incrementally widen the layers with $d$ branches by grouping the tasks based on the affinities of tasks, where $d$ is the number of clusters. The authors defined $A$ to be the affinity matrix, $i$, $j$ the task index, $E$ the expectation, and $k$, $l$ the branch index.
The error margin is defined
as $m^n_i = |t^n_i -s^n_i|$, where $t^n_i $ is the binary label for task $i$ at example $n$ and $s^n_i$ is the prediction.
The affinity of each pair of tasks $i,j$ is defined as $A(i,j) = P(e^n_i=1, e^n_j=1) + P(e^n_i=0, e^n_j=0) = E\{e^n_ie^n_j + (1-e^n_i)(1-e^n_j)\}$,
where $e^n_i$ is an indicator variable for task $i$ at example $n$. The indicator variable is set to be $1$ if $m^n_i$ is greater than the average error margin $E\{m_i\}$. To compute the affinity of two branches $k,l$ connecting to the current layer, the authors denoted $i_k$ and $j_l$ as the $i^{th}$ and $j^{th}$ tasks in $k$ and $l$ branches respectively. The affinity of two branches is defined by $A(k,l) = mean_{ik}(\min_{jl} A(i_k,j_l))$ and $A(l,k) = mean_{jl}(\min_{ik} A(i_k,j_l))$.
The final branch affinity score $A^f$ is the average of two affinities $k$ and $l$:

\begin{equation}
    \label{e3.1}
    A^f(k,l)  = (A(k,l)+A(l,k))/2.
\end{equation}
After getting the affinity matrix $A^f$, the authors performed spectral clustering to obtain a grouping function $g_d: [c] \rightarrow [d]$, which means $c$ old branches can be assigned to $d$ clusters. In order to determine the optimal number of branches, the authors minimized the following loss function:
\begin{equation}
    \begin{split} \label{e3.2}
        L^l(g_d) = (d-1)L_02^{p_t} + \alpha L_s(g_d)
    \end{split}
\end{equation}
where the first part is a penalty term for creating branches at layer $l$, the second part is the penalty for separation defined as: $L^i_s(g_d)  = 1 - mean_{k\in g^{-1}(i)}(\min_{l\in g^{-1}(i)} A_f(k,l))$ and $\alpha$ is a positive parameter.
% \he{How about $\alpha$?Done}
In our proposed method, we use the same method to approximate the similarities of tasks, but we target the more complex scenario with multiple views instead of a single view. In addition, different from the the fully adaptive layer-wise widening model whose input data is limited to a single modality (i.e., image data), our model is able to handle multi-modality data, such as text, image, video, etc.

\begin{figure*}[t]
\begin{tabular}{cc}
\hspace{-0.5cm}
\includegraphics[width=0.5\linewidth]{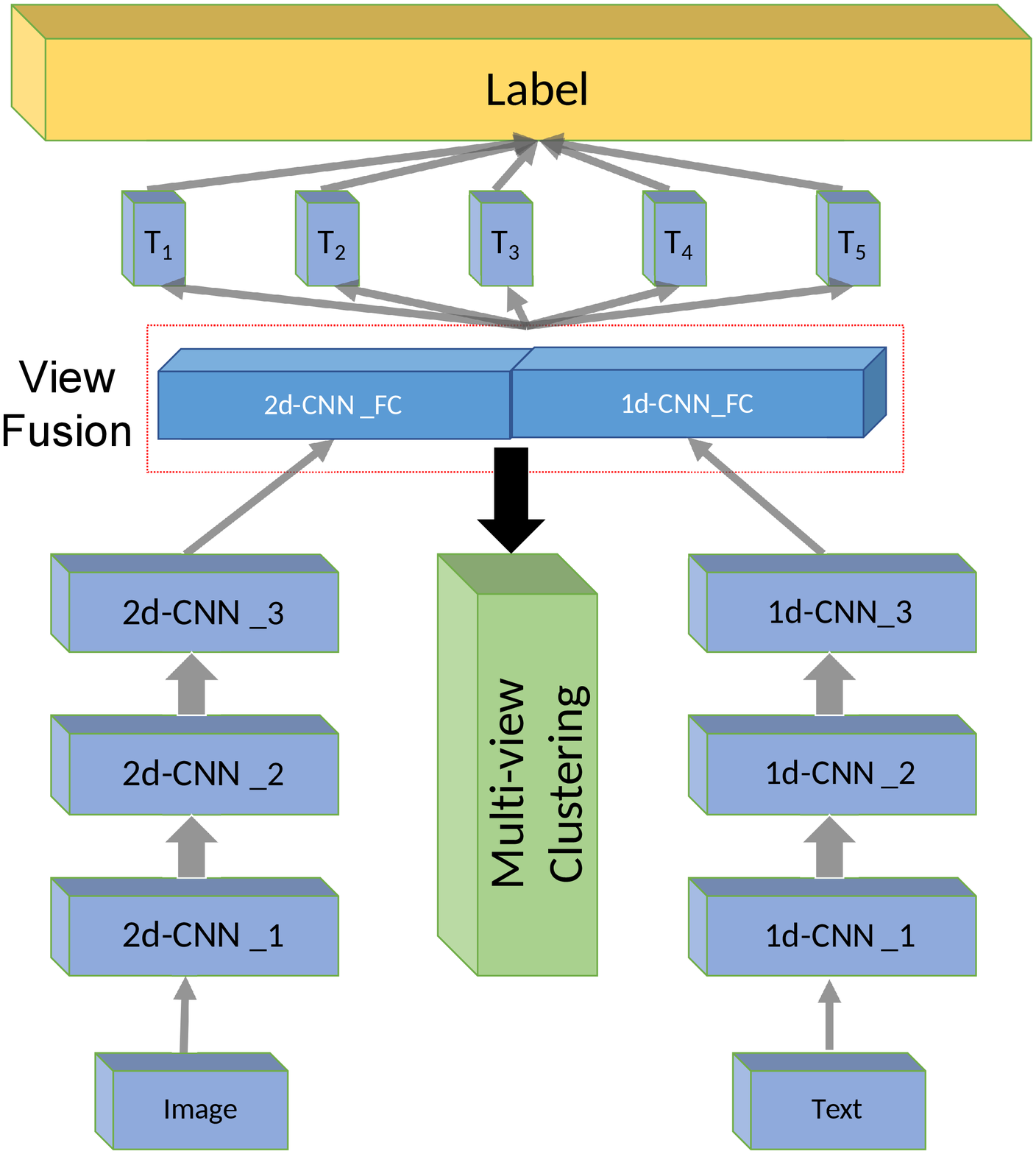} &
\includegraphics[width=0.5\linewidth]{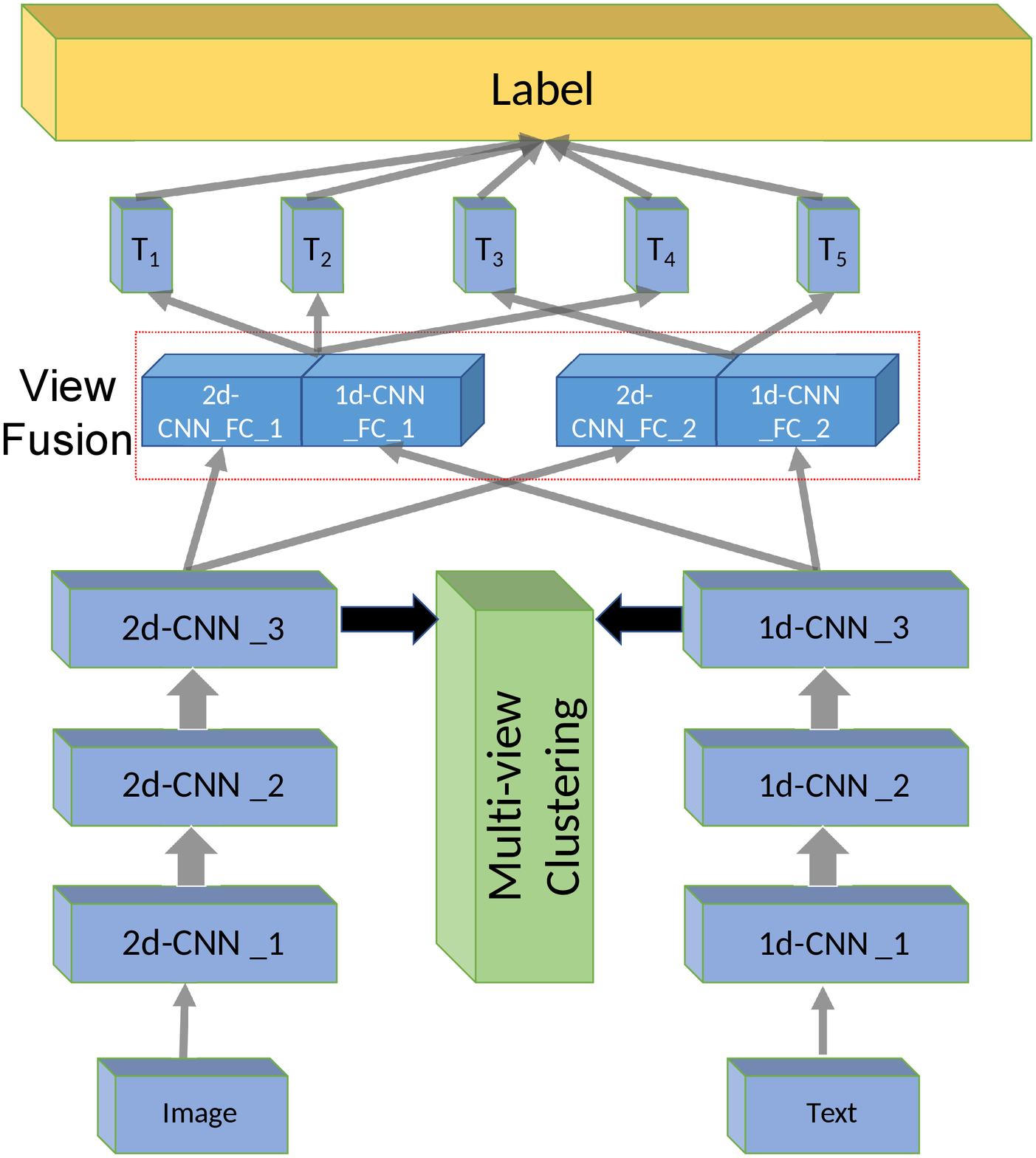}\\
(a) Round 1 & (b) Round 2\\
\end{tabular}

\caption{Suppose we are provided an image and text mixed data set with five tasks. We construct a 2d convolutional neural network (2d CNN) for image data and 1d CNN (or LSTM) for text data. In round 1, we aim to find the relatedness of tasks and multi-view clustering method is applied to decide how many branches we need to create and how to group tasks. After we train two neural networks with two types of data, we compute the similarities of tasks and update the structure in the next round. At the beginning of round 2, we decompose the split layer 2d-CNN$\_$FC into 2d-CNN$\_$FC$\_$1 and 2d-CNN$\_$FC$\_$2, and the split layer 1d-CNN$\_$FC into 1d-CNN$\_$FC$\_$1, 1d-CNN$\_$FC$\_$2, respectively. The size of 2d-CNN$\_$FC$\_$1, 2d-CNN$\_$FC$\_$2 are still the same as the size of 2d-CNN$\_$FC. Then, we assign tasks $T_1, T_2,T_4$ to 2d-CNN$\_$FC$\_$1 and 1d-CNN$\_$FC$\_$1, and tasks $T_3, T_5$ to 2d-CNN$\_$FC$\_$2 and 1d-CNN$\_$FC$\_$2 based on the clustering results. The filters or the kernels at the newly created branches (2d-CNN$\_$FC$\_$2 and 1d-CNN$\_$FC$\_$2) are initialized by directly copying from 2d-CNN$\_$FC or 1d-CNN$\_$FC. Next, we aim to find the similarities of the branches (the first branch: 2d-CNN$\_$FC$\_$1 and 1d-CNN$\_$FC$\_$1, the second branch: 2d-CNN$\_$FC$\_$2 and 1d-CNN$\_$FC$\_$2) and split 2d-CNN$\_$3 and 1d-CNN$\_$3 to create more branches by repeating these procedures.}
% \he{Did you introduce the two rounds in the main text at all?Done}
\label{fig1}
% \vspace{-0.5cm}
\end{figure*}

\subsection{Deep MTMV}
Now we are ready to introduce our proposed framework. The main idea of the proposed model is to utilize the label information from the training data as well as the consistency among different views to help classify the test data.
Suppose that the data set has $m$ views and $T$ tasks. We denote $\mathcal{D_L} = \{X_i^j, Y^j\}$ as the training data set, where $X_i^j \in R^{n_j \times d_i}$ corresponds to the feature matrix of the $i^{th}$ view and the $j^{th}$ task, $Y^j \in R^{n_j \times 1}$ consists of the class label of the $j^{th}$ task, $n_j$ is sample size of the $j^{th}$ task and $d_i$ is the dimensionality of the feature space in the $i^{th}$ view.
We denote $H$ to be a feature mapping of a neural network and $H_i(W_i,X_i^j,b_i)$ to be the output of the $i^{\rm th}$ view shared by $T$ tasks at the beginning with weights $W_i$ and biases $b_i$. To combine the feature mappings of multiple views, we denote $F$ to be the view fusion of multiple neural networks, which consists of several fully connected layers, and $F(\forall i, H_i(W_i,X_i^j,b_i)$\footnote{we slightly abuse the notation and it will be explained in later section.} outputs the label in the last fully connected layer (Details will be discussed in the Section 3.3). Figure \ref{fig1} provides the architecture of the framework, where we assume that the input consists of both image data and text data, although the proposed framework can be naturally generalized to handle additional data types. In this case, we construct two neural networks for two types of data, and $H_1(W_1,X_1^j,b_1)$ and $H_2(W_2,X_2^j,b_2)$ are the feature mappings of 2d CNN and 1d CNN (or Long Short-Term Memory, namely LSTM), respectively. $F(\forall i, H_i(W_i,X_i^j,b_i))$ combines the output of $H_1(W_1,X_1^j,b_1)$ and $H_2(W_2,X_2^j,b_2)$, and outputs the label.
Generally speaking, suppose that the data set has $m$ views and $T$ tasks.

The cost function of our algorithm can be written as:
\begin{equation} \label{e3.3}
    \begin{split}
        J & = \frac{1}{2}\sum_j^T\norm{F_j(\forall i, H_i(W_i,X_i^j,b_i))-Y_j}^2_2 + \sum_{i=1}^m \lambda_i\norm{W_i}_2 \\
    \end{split}
\end{equation}
where
$\lambda_i$ is a positive parameter. The objective of the proposed model is to fuse multiple views together, and to learn the weights of different views  automatically. The relatedness of tasks is exploited to improve the performance by applying the multi-view clustering method.

 \subsection{Regularization Layer}
 In multi-view learning, co-training~\cite{blum1998combining} is a commonly used method by utilizing the consistency principle to maximize the mutual agreement of label predicted by distinct views. In co-training, one uses multiple classifiers to predict the labels of the unlabeled data, adds the top $k$ confident unlabeled data into the training data set and repeats this procedure until all the unlabeled data has been added into the training set. However, in some situations, we cannot assume that each view has equivalent contribution to the prediction.
 Take web classification problem as an example, multiple views in this classification problem are the content of the web page, the title of the web page and the link within the web page. Obviously, the content of the web page contributes most to the prediction, while the title of the web page and the link within the web page have less contribution to the prediction. If the prediction made by the classifier trained on either the title of the web page or the link within the web page is used to label the unlabeled data, the prediction may not be as accurate as the prediction made by the classifier trained on the content of the web page. Therefore, assuming the equivalent contribution may result in a worse performance in some scenarios like this.

 To overcome this issue, we proposed multi-view fusion to automatically learn the weight of each view that contributes to the prediction. Given $m$ views, we have
 \begin{equation} \label{e3.4}
    F(H) = \sigma(W_F,H,b_F)\\
\end{equation}
 where $H = H_1(W_1,X_1^j,b_1)\oplus H_2(W_2,X_2^j,b_2)...\oplus H_m(W_m,X_m^j,b_m)$, the symbol $\oplus$ means concatenation, $\sigma(\cdot)$ is an activation function, and $W_F$ and $b_F$ are the weight and bias, respectively. In this equation, $H_i(W_i,X_i^j,b_i)$ can be considered as a feature mapping of the $i^{th}$ view and the expectation of the weights determines how many percentages each view contributes to the prediction of the training data. These weights are also used to determine which view is the centroid view when we apply the centroid multi-view clustering method to group similar tasks in the next subsection.

% \vspace{-0.1cm}
\subsection{Layer Widening and Task Clustering} %Multi-view Clustering
Similar to the structure proposed in~\cite{lu2016fully}, our training algorithm consists of a procedure to widen the layers of neural networks in order to explore the relatedness of multiple tasks and to group similar tasks into the same clusters based on the similarity among tasks. Most deep models assume that the tasks share the same parameters at the first several layers and have their own parameters at the following layers~\cite{zhang2017survey}. In our case, we have multiple neural networks to start with, each of which is associated with a single view. Furthermore, it is usually the case that each neural network ends up with its own distinct structure when we update its structure separately by grouping tasks based on their similarities within a single view. For example, suppose that both task A and task B are assigned to the same cluster in the first neural network (i.e., the first view), but in the second neural network (i.e., the second view), task A and task B may be assigned to two different clusters. This phenomenon contradicts our intuition that these neural networks created by multiple views should have the identical layer structure.

To address this problem, motivated by~\cite{misra2016cross}, we propose to insert a cross-stitch network into our architecture, in order to learn task relatedness inside the hidden layers of the multiple neural networks for multiple views. In this way, we obtain a unified task grouping informed by multiple views instead of potentially inconsistent groupings from different views. More specifically,
in our proposed model, we consider each task as a data point.
After the training stage, we compute the task-similarity matrices and estimate the affinity of the tasks in $m$ views according to equation \ref{e3.1}. To update the structure of neural networks, we create $d$ branches for a unit layer based on equation~\ref{e3.2}, initialize the weight of these newly created branches by directly copying from the split layer, and link the old branches from the previous layer to these newly created branches based on the result of the following clustering method. Because the weight of one view might be higher than other views, centroid-based co-regularized multi-view spectral clustering approach~\cite{kumar2011co} is used to assign similar tasks to the same group and dissimilar tasks to different groups in each round.
The intuition is that the underlying clustering would assign the corresponding task in each view to the same cluster~\cite{kumar2011co}. Given $m$ views, $T$ tasks, and $k$ clusters, we have
\begin{equation} \label{e3.5}
    \begin{split}
        & \max_{U_i}  \sum_{i=1}^m (U_i' L_i U_i) + \sum_{i=1}^m \lambda_i tr( U_iU_i'U_cU_c') \\
        & s.t \hspace{0.3cm}  U_i'U_i = I, 1 \leq i \leq m,U_c'U_c = I
    \end{split}
\end{equation}
where $L_i \in R^{T \times T}$  is the graph Laplacian of the $i^{th}$ view , $U_i \in R^{T \times k}$ consists of the eigenvectors of the $i^{th}$ view, $U_c \in R^{T \times k}$ consists of the eigenvectors from the most important view and $\lambda_i$ is the weight of the $i^{th}$ view. The normalized graph Laplacian $L_i$ of the $i^{th}$ view is defined as $L_i=D^{-\frac{1}{2}}A_i^fD^{-\frac{1}{2}}$, where $A^f_i$ is the affinity matrix for the $i^{th}$ view based on equation \ref{e3.1} and $D \in R^{T \times T}$ is diagonal matrix with $D_{\alpha,\alpha}$ to be the sum of the $\alpha^{th}$ row of $A^f_i$. The detailed approach to solve this optimal problem can be found in ~\cite{kumar2011co}, and is omitted here for brevity. The optimal solution determines how tasks are clustered and how a layer is split.

In addition, we can naturally extend this multi-view clustering method to accommodate the scenario where some views may be missing for some tasks as in ~\cite{zhou2007spectral}.  Although for the missing views, the corresponding entries of some tasks in the affinity matrices would be unavailable, these missing similarities can be estimated by the corresponding similarities in other views. In Section 3.3, our model automatically learns the weights of different views, with which the missing entries of the affinity matrices can be approximated by averaging the similarities from the available views. Besides, the learned weights of different views can also be used to set the parameters $\lambda_i$ during the multi-view clustering process.

\begin{algorithm}[t]

    \textbf{Input:}{ The initialized model $M$, the total number of round $R$, the training data set $\mathcal{D_L} =\{X_i^j, Y^j\}$ and the number of branches $b$.}\\
    \textbf{Output:}{ The well-trained model $M$.\\}
    \textbf{Initialization}: Load pre-trained model or randomly initialize the weights and biases of the model, and set t to be 0.
    % \he{You should initialize $t$ as well.Done}

    \While{$t \leq R$ and $b > 1$}{
        \textbf{Step 1: }Train the model $M$ with training data $\mathcal{D_L}$.

        \textbf{Step 2: }Compute the affinity matrices about the tasks similarities for $m$ views based on equation~\ref{e3.1}.

        \textbf{Step 3: }Determine the number of clusters by multi-view clustering method based on equation~\ref{e3.2} and ~\ref{e3.5}.

        \textbf{Step 4: }Create branches and widen layers for $M$ based on the results of multi-view clustering.

        \textbf{Step 5: }$b$ $\gets$ the number of branches in the current layer.

        \textbf{Step 6: }$t$ $\gets $ $t + 1$.

        }
    Train the model $M$ until convergence.
    % Predict the label of test data $Y_U$.
    \caption{Deep-MTMV}
    \label{alg1}
    % \vspace{-0.1cm}
\end{algorithm}
% \vspace{-0.1cm}

\subsection{Multimodality Model for text and image mixed data set}
As mentioned before, our proposed framework is able to take as input multimodality data. Next, we use text and image mixed data set to illustrate the key idea. Given two sources of data: image data and text data, we build a convolutional neural network for image data, and a 1d convolutional neural network~\cite{kim2014convolutional} (or Long Short-Term Memory) for text data. Notice that the specific choice of the neural network for each data modality is orthogonal to the proposed framework. Furthermore, the text data is pre-processed by word2vec algorithm~\cite{mikolov2013distributed} to extract word embeddings as the input of 1d CNN. The vital features, such as unigram, bi-gram, and tri-gram, are extracted by the filters of CNN with different size, such as $K_1 \in R^{1 \times d}$, $K_2 \in R^{2 \times d}$, and $K_3 \in R^{3 \times d}$, where $d$ is the dimensionality of word2vec embeddings. At first, two neural networks are trained separately, and then the feature mappings extracted by the two neural networks are appended in the fully connected layers to predict the labels of test data. In addition to the data sets containing the same data type, such as CelebA~\cite{liu2015deep}, WebKB, we will present experimental results on a real-world data set, FamousFood, which contains two types of data, image and text, to evaluate the performance of our proposed framework.

% \vspace{-0.1cm}
\subsection{The Proposed Algorithm}
Our proposed algorithm is presented in Algorithm \ref{alg1}. It takes a initialized model (which is obtained based on the initialization algorithm in ~\cite{lu2016fully}
% \he{Please fill in the details.Done})
, training data, the number of branches, and the total number of rounds as inputs, and outputs the well-trained model. The algorithm works as follows. We first construct a neural network structure for each view at the beginning, train $m$ neural networks with the training data after initialization and fuse these neural networks in the regularization layer to get a final model. Then, we compute the affinity matrices about the tasks similarities. After the number of clusters is determined by minimizing the loss of the multi-view clustering method, we create new branches and assign the similar tasks into the same branch and dissimilar tasks into the different branches. When the number of round reaches its maximal or the branches cannot be split, then stop updating the structure and train the model until convergence.

\section{Experimental Results}
In this section, we demonstrate the performance of our proposed Deep-MTMV algorithm in terms of effectiveness by comparing with state-of-the-art methods.

\begin{figure*}[t]% config front 60; legend 40; x axis 4.6
\begin{center}
\begin{tabular}{cc}
\hspace{-0.5cm}
\includegraphics[width=0.5\linewidth]{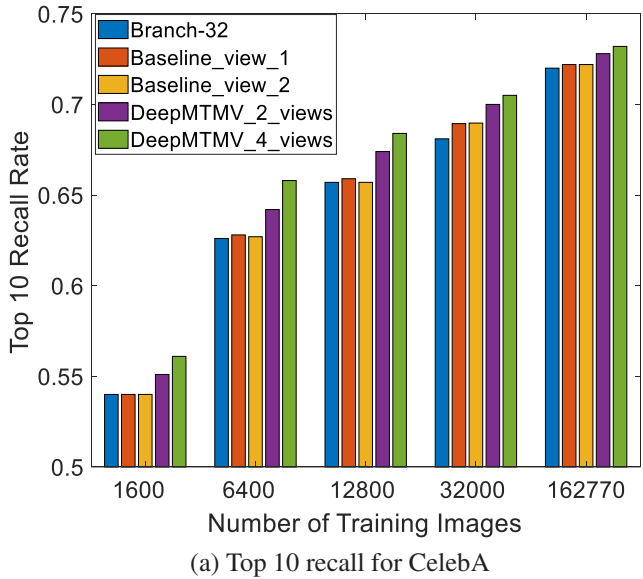} &
\includegraphics[width=0.5\linewidth]{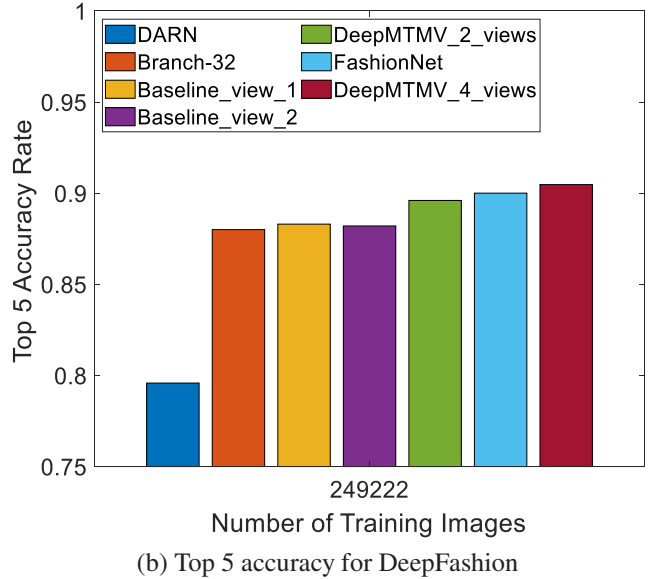}\\
% \vspace{-0.5cm}
(a) Top 10 recall for CelebA & (b) Top 5 accuracy for DeepFashion\\
\end{tabular}
\end{center}

\caption{Effectiveness Analysis (Best viewed in color)}
% \he{The size of the legend is not consistent.Done}}
\label{fig2}

\end{figure*}

\subsection{Data sets} In this paper, we evaluate our proposed algorithm on the following data sets:
\begin{itemize}
    \item \textbf{CelebA~\cite{liu2015deep}}: It is composed of 202,599 images of celebrities, and 40 labeled facial attributes. Each attribute, such as black eye, brown eye, bald, is considered as one task in this classification problem. We extract two views or four views from each image in a way mentioned below. In our setting, we have 40 different tasks for 40 attributes and 2 (or 4) views.
    % \he{More details like FamousFood.Done}
    \item \textbf{Deepfashion~\cite{liu2016deepfashion}}: It consists of 50 categories and more than 289,222 images of clothes. Each category, such as hoodie and ramper, is considered as one task in this classification problem. We extract two views or four views from each image in a way mentioned below. In this setting, we have 50 different tasks for 50 different categories and 2 (or 4) views.
    % \he{More details like FamousFood.Done}
    \item \textbf{WebKB}: This is a textual data set, which consists of over 4000 web pages from 4 universities and includes 3 views,
    % \he{The previous sentence has grammar mistakes.Done}
    including the content of the web page, the title of the web page and the links within the web page. In our setting, each university is treated as a task and our goal is to classify each web page as course or non-course.
    \item \textbf{FamousFood}: In this data set, the images of famous food and the text of food description are crawled from the online photo sharing website Flickr. This data set contains 4 types of foods, which fall into 2 categories (sweet food or fast food), and each is considered as a task in our setting. Two different data source are image and the related text. For each food, it contains more than 450 images on average. In our setting, we have 6 (4 types of foods and 2 food categories) tasks and 2 views.
    % \yuc{Need detailed information on WebKB and Caltech.}
\end{itemize}

\subsection{View extraction for two image data sets (CelebA and Deepfashion)} In our experiments, we extract two views from a single image by splitting the width of each single image into two sets of indices: the even indices and the odd indices. Keeping the height of each image unchanged and combining all odd (even) indices together, we form the first (second) view. The way to get the four views is similar to the way to get the two views. Instead of only splitting the width of a single image into four views, we divide the single image into two parts both vertically and horizontally. By selecting the odd indices of the width and the odd indices of the height, we get the first view (and we can get the other three views in a similar way). The reason why we split the image this way is that we want to keep the views from overlapping.

\begin{table}
\centering
\hspace{0.2cm}
\begin{tabular}{|m{2.7cm}|m{2.7cm}|m{1.2cm}|}
\hline \textbf{Pre-module}       & \textbf{Post-module} & \textbf{P value}   \\
\hline Branch-32                 & Deep-MTMV\_4\_views  &  1.01E-17          \\
\hline Baseline\_view\_1         & Deep-MTMV\_4\_views  &  4.33E-56          \\
\hline Baseline\_view\_2         & Deep-MTMV\_4\_views  &  9.24E-61          \\
\hline Deep-MTMV\_2\_views       & Deep-MTMV\_4\_views  &  8.66E-10          \\ \hline
\end{tabular}
\caption{Student T Test with 95\% Confidence Level}
\label{table1}
\end{table}
\vspace{-0.2cm}

\begin{figure*}[t]
\begin{center}
\begin{tabular}{cc}
% \vspace{-0.1cm}
\hspace{-0.5cm}
\includegraphics[width=0.5\linewidth]{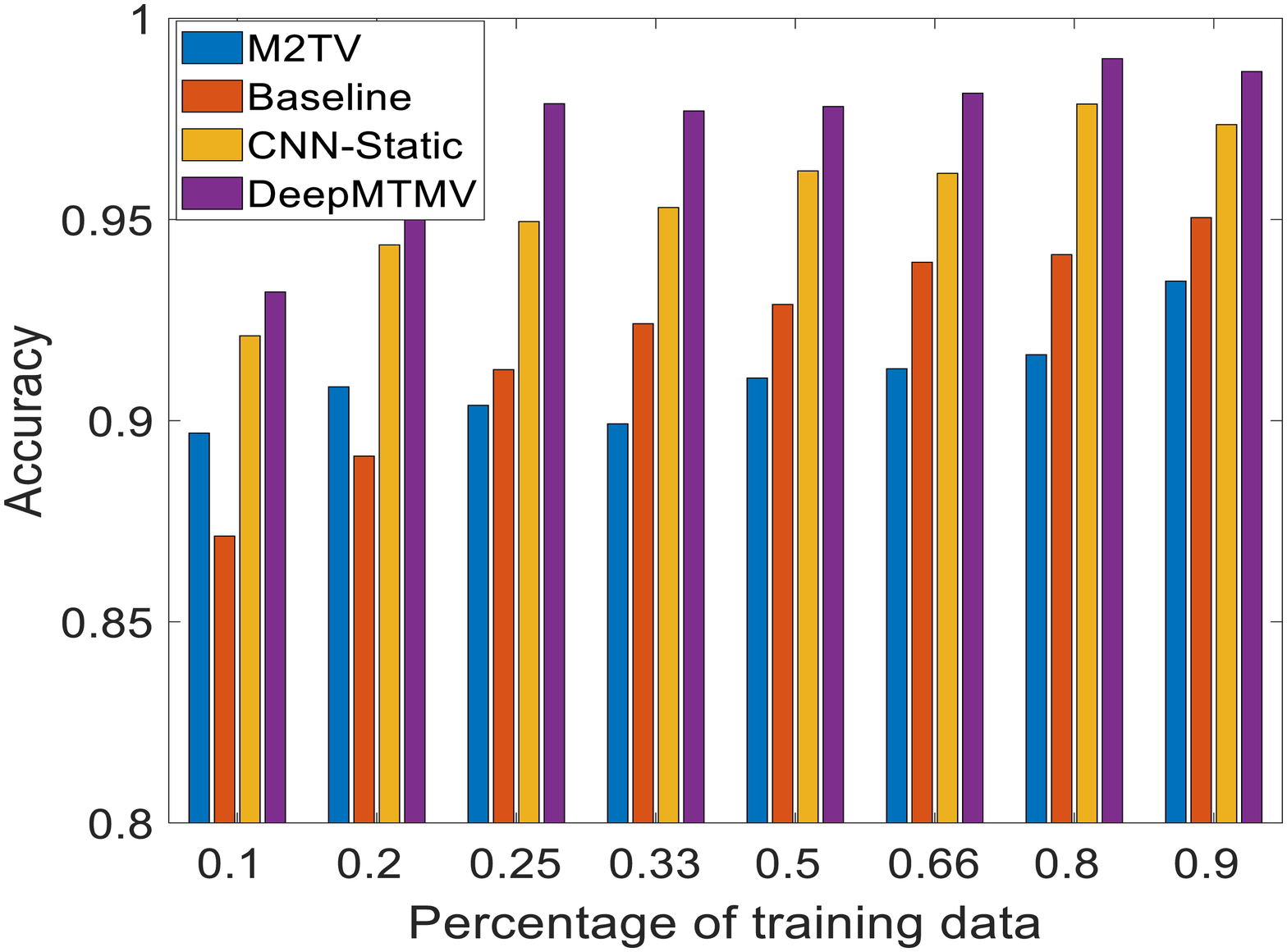} &
\includegraphics[width=0.5\linewidth]{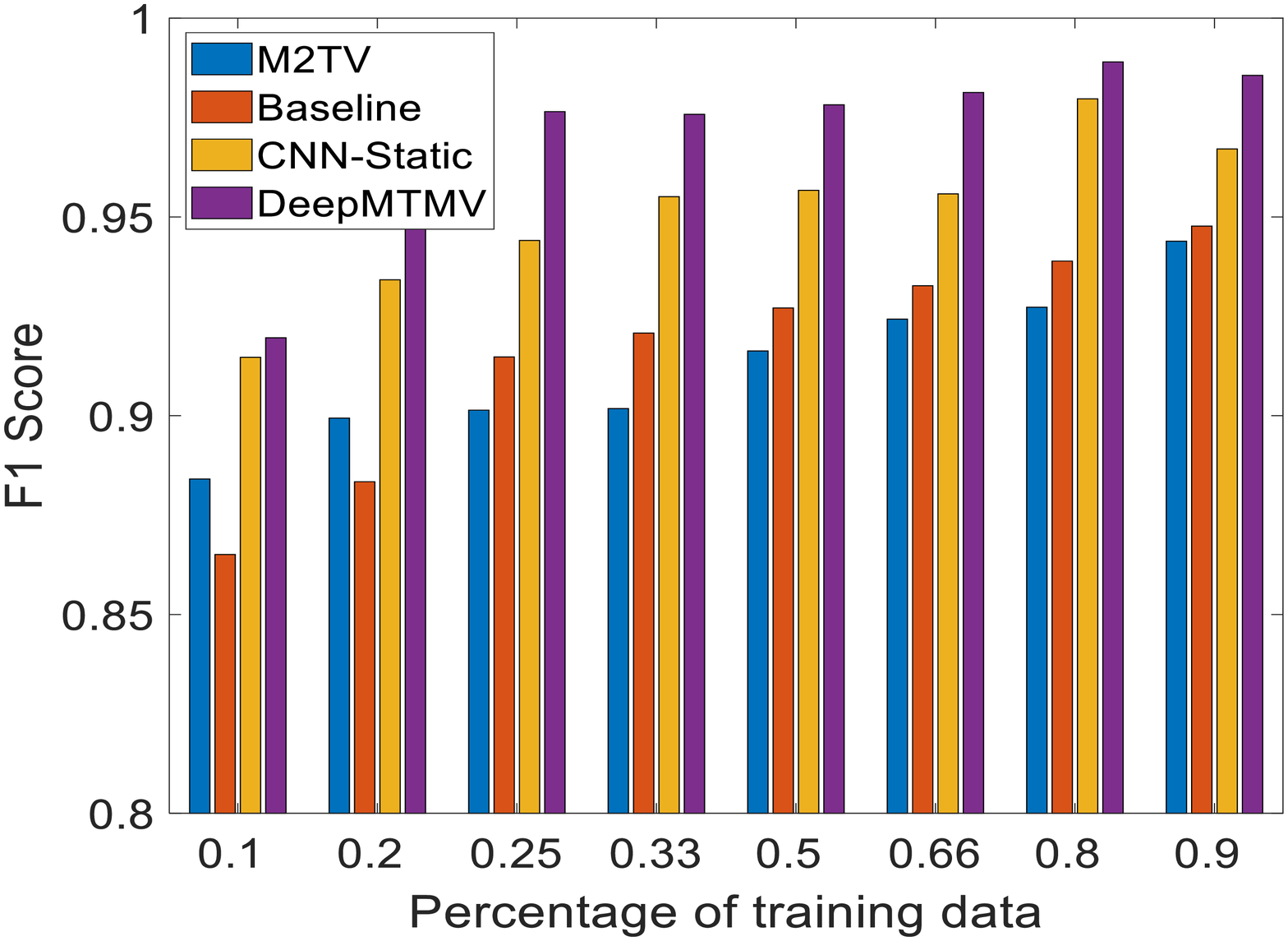}\\
\vspace{-0.5cm}
(a) Accuracy for WebKB & (b) F1 score for WebKB\\
\end{tabular}
\end{center}
\caption{Results of WebKB (Best viewed in color)}
\label{fig3}

\end{figure*}

% \subsection{Comparison Results}
\subsection{Comparison methods} In our experiments, we compare the performance of the following methods:
(1). Our baseline model trained with view 1;
(2). Our baseline model trained with view 2;
(3). Branch-32~\cite{lu2016fully} (CelebA, DeepFashion and FamousFood data set);
(4). FashionNet ~\cite{liu2016deepfashion} (DeepFashion data set);
(5). DARN ~\cite{huang2015cross} (DeepFashion data set);
(6). M2TV~\cite{he2011graphbased} (WebKB data set and FamousFood data set);
(7). CNN-Static \cite{kim2014convolutional} (WebKB data set).

Notice that some of these methods are best suited for certain data modalities. For example, Branch-32, FashionNet are designed to deal with image data; and M2TV, CNN-Static are only good at coping with text data. Therefore, we omit the results of these methods on the non-applicable data sets.

\subsection{Two images data sets: CelebA and Deepfashion}
The labels of two images data set are attributes of object or person in the image.  The comparison results conducted on the two data sets in terms of the top 10 recall rate and top 5 accuracy rate metrics are shown in Figure~\ref{fig2} (a) and~\ref{fig2} (b) respectively. In both figures, the x-axis is the number of training images. The y-axis in Figure~\ref{fig2} (a) is the top 10 recall rate for CelebA data set and the y-axis in Figure~\ref{fig2} (b) is the top 5 accuracy rate for DeepFashion data set. These two figures show that our algorithm outperforms the others with respect to these two evaluation metrics. From the two figures, we can observe that the results of Deep-MTMV with four views outperforms Deep-MTMV with two views. Our intuition of this observation is that four views of Deep-MTMV model can preserve more spatial information than Deep-MTMV model with two views.
Moreover, to further prove that our proposed algorithm leads to significant improvement, we conduct the paired  student t test, which is shown in Table~\ref{table1}.
% For example,
We compared our methods with Branch-32 method and we found that the $p$ value is 1.01E-17, which indicates that our model does lead to significant improvements over other methods on average. In addition, the $p$ value of the paired student t test on Deep-MTMV with two views and Deep-MTMV with four views indicates that the more views we have, the more spatial information of image we can preserve, and thus a better performance.

\subsection{Text data set: Webkb}
Next, we test the performance of our proposed model on WebKB data set, and the goal is to classify each web page as course or non-course.
The baseline method is a simple version of our proposed method, which is trained on a single view, i.e., the content of web page. To test the performance of CNN-Static, we concatenate three views together to be the input of this model. The comparison results in terms of the accuracy and the F1 score are shown in Figure~\ref{fig3} (a) and Figure~\ref{fig3} (b), respectively. The x-axis in these two figures is the percentage of training data and the y-axis is accuracy in Figure~\ref{fig3} (a) and F1 score in Figure~\ref{fig3} (b), respectively. These two figures show that our proposed model is better than the others with respect to both evaluation metrics. From these figures, we observed that the accuracy and F1 score of our proposed model can be as high as 92\%, even if only 10\% of training data is provided. When more than 80\% of training data is given, the accuracy rate and F1 score reach 99\%. In this experiment, we also evaluate the weight of each view that contributes to the final prediction. After the model is well-trained, the weight of the content of web page is around 0.0561, compared with 0.0501 and 0.0498 for the rest two views, which is consistent with our expectation that the web content is a little bit important than the link and the title.

\subsection{Text and image mixed data set: FamousFood}
Finally, we evaluate our model on text and image mixed data set. Due to the limitation of the compared models, Branch-32 only trains on the single view (image); to reduce the feature dimension for M2TV, the size of each image is re-sized from 224x224 to 50x50 and then each image is converted to a vector. The comparison results in terms of the accuracy and the F1 score are shown in Table~\ref{table2}. We measure the accuracy of type of food prediction, the accuracy of food category prediction, and the macro F1 score. This table shows that our proposed model outperforms the others with respect to these evaluation metrics. From this table, we observed that the accuracy of food prediction reaches 71.22\% compared with 59.71\% achieved by Branch-32 and 48.75\% achieved by M2TV. The worse performance of M2TV for this data set might be due to the fact that M2TV fails to capture the spatial information of images, while Branch-32 cannot utilize the complementary information from another view to further improve the performance.

\begin{table}
\centering
\begin{tabular}{|m{2cm}|m{1.8cm}|m{1.8cm}|m{1.2cm}|}
\hline \textbf{Model}       & \textbf{Accuracy of food prediction} & \textbf{Accuracy of category prediction}    & \textbf{F1 score}\\
\hline Branch-32                 & 59.71\%  &  90.64\%    &  55.18\%      \\
\hline M2TV                      & 48.75\%  &  69.44\%    &  49.18\%      \\
\hline Deep-MTMV                 & 71.22\%  &  94.60\%    &  71.96\%      \\ \hline
\end{tabular}
\caption{Results for FamousFood data set}
\label{table2}
% \vspace{-0.5 cm}
\end{table} 
\section{Conclusion}
In this paper, we propose a deep multi-task multi-view learning framework, i.e., Deep-MTMV. It trains multiple neural networks, automatically learns the weight of the different views that contribute to the prediction in the regularization layer, groups similar tasks together based on the relatedness of tasks, and classifies the test data with a high accuracy. To the best of our knowledge, the proposed framework is the first deep model for jointly addressing task and view dual heterogeneity, particularly for a data set with multiple modalities. Furthermore, we generalize the proposed Deep-MTMV algorithm to solve multiple real image and text classification problems by (1) utilizing the complementary principle and the consensus principle of multiple views, and (2) learning the relatedness of tasks in each layer of the deep networks. Finally, we compare our algorithm with state-of-the-art techniques, and conduct experiments on multiple real-world data sets to demonstrate that our algorithm leads to statistically significant improvements in the performance. Applying our approach to other applications \cite{video} is one of the future work.

\section*{Acknowledgement}
This work is supported by the National Science Foundation under Grant No. IIS-1552654, Grant No. IIS-1813464 and Grant No. CNS-1629888, the U.S. Department of Homeland Security under Grant Award Number 17STQAC00001-02-00, and an IBM Faculty Award. The views and conclusions are those of the authors and should not be interpreted as representing the official policies of the funding agencies or the government.

\bibliographystyle{abbrv}
{\small
\bibliography{references}}

\end{document}